\documentclass[letterpaper, 10 pt, conference]{ieeeconf}  %

\IEEEoverridecommandlockouts                              %

\usepackage{cite}
\usepackage{amsmath,amssymb,amsfonts}
\usepackage{algorithmic}
\usepackage{graphicx}
\usepackage{textcomp}
\usepackage{xcolor}
\usepackage{eso-pic}

\def\BibTeX{{\rm B\kern-.05em{\sc i\kern-.025em b}\kern-.08em
    T\kern-.1667em\lower.7ex\hbox{E}\kern-.125emX}}

\usepackage{tikz}
\usepackage{pgfplots}
\usepackage[per-mode=fraction, detect-weight=true]{siunitx}
\usepackage[hidelinks]{hyperref}
\usepackage{algorithm}
\usepackage{lipsum}
\usepackage[inkscapelatex=false, inkscapepath=./build/svg-inkscape]{svg}
\usepackage{acronym}

\DeclareSIUnit\bar{bar}
\usepgfplotslibrary{groupplots}
\usetikzlibrary{backgrounds}
\usetikzlibrary{plotmarks}
\usetikzlibrary{fit}
\pgfplotsset{
	compat=1.18,
	standard-plot/.style={
			grid=major,
			grid style={dashdotted},
		},
	legend style={
			font=\fontsize{7}{11}\selectfont
		},
label style={font=\fontsize{7}{11}\selectfont},
	line-plot/.style={
			width=\columnwidth,
			height=0.5\columnwidth,
		},
	xy-plot/.style={
			width=\columnwidth,
			height=0.5\columnwidth,
			standard-plot,
		},
	group-plot/.style={
			width=\columnwidth,
			height=0.39\columnwidth,
		},
	group-plot-2/.style={
			group-plot,
			group style={
					group size=1 by 2,
					xlabels at=edge bottom,
					xticklabels at=edge bottom,
					vertical sep=0cm,
				},
		},
	group-plot-3/.style={
			group-plot,
			group style={
					group size=1 by 3,
					xlabels at=edge bottom,
					xticklabels at=edge bottom,
					vertical sep=0cm,
				},
		},
	group-plot-4/.style={
			group-plot,
			group style={group size=1 by 4,
					xlabels at=edge bottom,
					xticklabels at=edge bottom,
					vertical sep=0cm,
				},
		},
	group-plot-5/.style={
			group-plot,
			group style={group size=1 by 5,
					xlabels at=edge bottom,
					xticklabels at=edge bottom,
					vertical sep=0cm,
				},
		},
	first-group-plot/.style={
			standard-plot,
			enlarge x limits=0,
			xmin=0,
			y label style={yshift=-0.6em},
		},
	middle-group-plot/.style={
			standard-plot,
			enlarge x limits=0,
			xmin=0,
			y label style={yshift=-0.6em},
		},
	last-group-plot/.style={
			standard-plot,
xlabel=Time (\si{\second}),
			enlarge x limits=0,
			xmin=0,
			y label style={yshift=-0.6em},
		},
	generic-linestyle/.style={thick},
	line-1/.style={generic-linestyle, TUMBlack},
	line-2/.style={generic-linestyle, TUMOrange},
	line-3/.style={generic-linestyle, TUMBlue},
	line-4/.style={generic-linestyle, TUMGreen},
	multiline-1/.style={line-1, mark=x, mark repeat=160},
	multiline-2/.style={line-2, mark=o, mark repeat=160, mark phase = 40},
	multiline-3/.style={line-3, mark=triangle, mark repeat=160, mark phase=80},
	multiline-4/.style={line-4, mark=square, mark repeat=160, mark phase=120},
	background-1/.style={thin, TUMBlack!50},
	background-2/.style={thin, TUMOrange!50},
	background-3/.style={thin, TUMBlue!50},
	background-4/.style={thin, TUMGreen!50},
	line-max/.style={generic-linestyle, TUMOrange, dashed},
	line-actual/.style={generic-linestyle, TUMBlue},
}
\usetikzlibrary{shapes}
\usetikzlibrary{positioning}
\usetikzlibrary{pgfplots.statistics}

\tikzset{
architecture/.style = {
standard-block/.style={
		inner xsep = 0cm,
		align=center,
		font=\small
	},
software-block-circle/.style={
		standard-block,
		minimum height=1.5cm,
		circle,
		white,
		draw=TUMBlue,
		fill=TUMBlue,
	},
software-block/.style={
		standard-block,
		rectangle,
		rounded corners=.2cm,
text width=1.65cm,
		minimum height=1cm,
		minimum width=1.65cm,
		white,
		draw=TUMBlue,
		fill=TUMBlue,
	},
standard-arrow/.style={-latex},
standard-arrow-label/.style={fill=none, align=center, text width =1.5cm, font=\fontsize{7}{11}\selectfont},
		}
}

\acrodef{esc}[ESC]{electronic stability control}
\acrodef{abs}[ABS]{anti-lock braking system}
\acrodef{tc}[TC]{traction control}
\acrodef{cs}[CS]{countersteer system}
\acrodef{sc}[SC]{slip control}
\acrodef{pid}[PID]{proportional-integral-derivative}
\acrodef{lqr}[LQR]{linear quadratic regulator}
\acrodef{mpc}[MPC]{model predictive control}
\acrodef{pmpc}[PMPC]{pointmass model predictive control}
\acused{pmpc}
\acrodef{nmpc}[NMPC]{nonlinear model predictive control}
\acused{nmpc} 
\definecolor{TUMBlue}{RGB}{0,101,189}%
\definecolor{TUMWhite}{RGB}{255,255,255}%
\definecolor{TUMBlack}{RGB}{0,0,0}%
\definecolor{TUMBlue1}{RGB}{0,51,89}%
\definecolor{TUMBlue2}{RGB}{0,82,147}%
\definecolor{TUMGray1}{RGB}{51,51,51}%
\definecolor{TUMGray2}{RGB}{127,127,127}%
\definecolor{TUMGray3}{RGB}{204,204,204}%
\definecolor{TUMBlue3}{RGB}{100,160,200}%
\definecolor{TUMBlue4}{RGB}{152,198,234}%
\definecolor{TUMIvory}{RGB}{218,215,203}%
\definecolor{TUMOrange}{RGB}{227,114,34}%
\definecolor{TUMGreen}{RGB}{162,173,0}%

\newlength{\nodewidth}
\newlength{\nodespacinghor}
\newlength{\nodespacingver}
\newlength{\arrowlength}
\newcommand{\inputtikzfig}[1]{
	\includegraphics{images/tikz/#1.pdf}
}

\newcommand{\inputtikzplot}[1]{
	\includegraphics{images/tikz/#1.pdf}
}
 
\title{\LARGE \bf
Stability Control for Real World Testing in Autonomous Racing
\thanks{$^{1}$Technical University of Munich, Germany; School of Engineering \& Design, Department of Engineering Physics and Computation, Institute of Automatic Control}
\thanks{$^{2}$Technical University of Munich, Germany; School of Engineering \& Design, Department of Mobility Systems Engineering, Institute of Automotive Technology\newline{}
	Corresponding author: \href{mailto:phillip.pitschi@tum.de}{phillip.pitschi@tum.de}}
}

\author{Phillip Pitschi$^{1}$, Simon Sagmeister$^{2}$, Frederik Werner$^{2}$, Markus Lienkamp$^{2}$, Boris Lohmann$^{1}$}%

\newcommand\copyrighttext{%
	\footnotesize \textcopyright 2026 IEEE.  Personal use of this material is permitted.  Permission from IEEE must be obtained for all other uses, in any current or future media, including reprinting/republishing this material for advertising or promotional purposes, creating new collective works, for resale or redistribution to servers or lists, or reuse of any copyrighted component of this work in other works.
}

\newcommand\copyrightnotice{%
	\AddToShipoutPictureFG*{%
		\AtPageLowerLeft{%
			\raisebox{25pt}[0pt][0pt]{%
				\makebox[\paperwidth][c]{%
					\fbox{\parbox{\dimexpr\textwidth-2\fboxsep-2\fboxrule\relax}{\copyrighttext}}%
				}%
			}%
		}%
	}%
}

\begin{document}

\bstctlcite{IEEEexample:BSTcontrol}

\maketitle
\thispagestyle{empty}
\pagestyle{empty}
\copyrightnotice{}

\setcounter{footnote}{1}
\begin{abstract}
    Controlling an autonomous vehicle at the limits of handling is a challenging task. 
    Due to external influences, such as road conditions or weather, a vehicle can easily become unstable. 
    Since most control algorithms assume stable vehicle behavior, they might fail in these situations. 
    Especially when operating expensive vehicles without a safety driver on board, as in autonomous racing, this poses a significant challenge. 
    To enable safe operation at the vehicle's dynamic limits, we present a comprehensive stability control system that safeguards motion control algorithms in autonomous driving. 
    The proposed system consists of an \ac{esc}, a \ac{sc}, and a \ac{cs}, which collectively adapt steering and brake commands from the motion controller to maintain vehicle stability. 
    We validate our approach through both simulation and experiments on a real-world, full-scale vehicle. 
    The results show that the stability control system maintains vehicle stability in critical situations and extends the operational feasible region. 
    To simplify integration, we provide an open-source implementation at \href{https://www.github.com/TUMFTM/tam-stability-control}{github.com/TUMFTM/tam-stability-control}.
\end{abstract}

\section{Introduction}

Autonomous driving is a fast-growing research field across various domains. However, only a few algorithms are tested in real-world full-scale deployments \cite{betz2022AutonomousVehiclesEdge}.
While simulation approaches are continuously improving, evaluating the performance and robustness of control algorithms in the real world remains challenging \cite{chen2023MilestonesAutonomousDrivinga}.
In real-world operation, the vehicle's dynamic response can vary drastically due to factors such as temperature, road surface contamination, and weather conditions, including rain and wind.
Since the vehicle's response is heavily influenced by these factors, operating a vehicle at its dynamic limits can frequently result in unstable behavior, such as oversteer \cite{weber2024ModelingControlDynamic}.
However, most state-of-the-art motion control algorithms are designed with stable vehicle behavior in mind \cite{coulter1992ImplementationPurePursuit,weber2024ModelingControlDynamic, zhao2022JustifyingEmergencyDrift}. Therefore, handling unstable oversteer situations is particularly challenging for these approaches \cite{zhao2022JustifyingEmergencyDrift}.
Without active countermeasures, the application of such algorithms could lead to a sudden loss of control.
While there are \acf{sc} and \acf{esc} algorithms \cite{bosch2022automotive} that mitigate these risks in passenger vehicles, they are not yet integrated with a \acf{cs} or autonomous driving controller.

\begin{figure}
    \centering
    \includegraphics[width=0.92\linewidth]{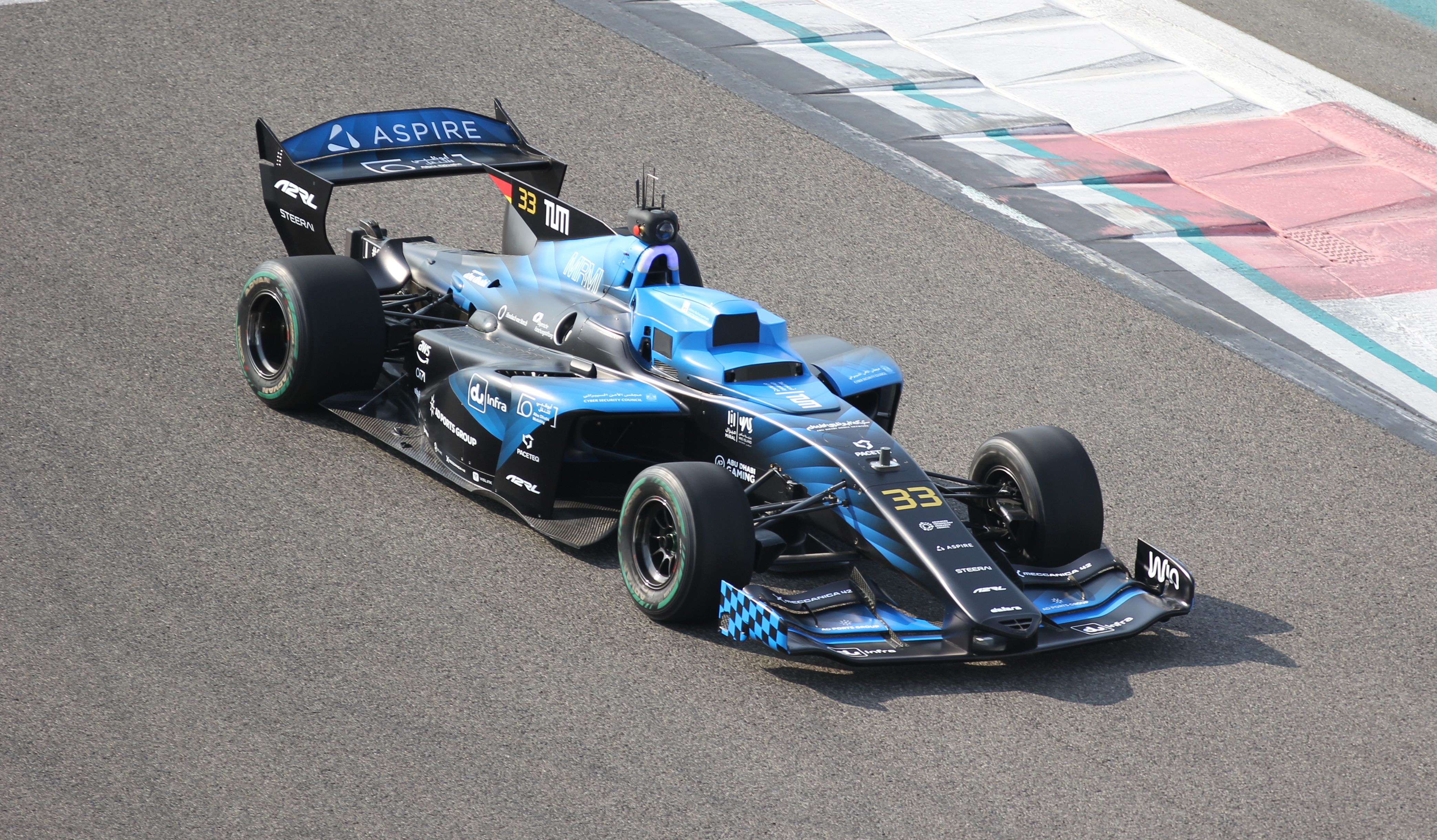}
    \caption{Dallara EAV25 race car of the TUM Autonomous Motorsport team at Yas Marina Circuit in Abu Dhabi.}
    \label{fig:hailey}
\end{figure}

Losing control of the vehicle poses a significant safety risk, leading to potentially costly crashes.
Therefore, maintaining vehicle stability is crucial for real-world testing of autonomous driving control algorithms, especially when operating at the vehicle's dynamic limits.
This is particularly important for autonomous race cars, which are often used as a testbed for developing and evaluating new control algorithms \cite{hoffmann2026HeadtoHeadautonomousracing}.
These autonomous race cars, such as the one shown in Fig.~\ref{fig:hailey}, are expensive and cannot be driven with a safety driver onboard. Furthermore, remote intervention is often slow due to transmission delays, which is why stability and crash-free operation of the autonomous system are essential. Even just locking the tires under braking or spinning the car without crashing can damage a set of tires, which poses a competitive disadvantage because the available sets are limited during a testing period.

To mitigate these risks, we introduce a stability control system for autonomous driving control algorithms. We combine subsystems for controlling longitudinal slip and lateral stability into a full stability control system that actuates both steering and wheel-individual brakes.
In summary, this paper comprises the following contributions:
\begin{itemize}
    \item We present a modular concept for a stability control system to safeguard algorithms for autonomous driving motion control, ensuring safe operation up to the vehicle's dynamic limits.
    \item We experimentally validate the presented approach in simulation and on a real-world vehicle. The results show that the stability control system can keep the vehicle stable in critical situations, including oversteer and excessive wheelslip.
    \item We demonstrate for three different motion control algorithms that our approach extends the vehicle's drivable lateral and longitudinal acceleration envelope, effectively improving lap times.
    \item We provide a modular and validated C++ implementation at \href{https://www.github.com/TUMFTM/tam-stability-control}{github.com/TUMFTM/tam-stability-control}.
\end{itemize}

\section{Related Work}

This section presents an overview of the state of the art in \acf{sc} and yaw rate control algorithms.

\ac{sc} algorithms are typically classified into rule-based \cite{bosch2022automotive, piccinini2025nnabs}, fuzzy logic \cite{cabrera2006fuzzyabs,lin2003fuzzylearningabs}, neural network-based \cite{Pedro2012}, \ac{pid} \cite{savitski2018robust}, \ac{lqr} \cite{petersen2003lqrabs}, sliding mode \cite{bhandari2012abswithsurfaceprediction}, and \ac{mpc} strategies \cite{pretagostini2020surveyabs, basrah2017mpcslipcontrol}. These approaches are frequently integrated with road friction estimation modules to accommodate varying road conditions. The estimation modules are implemented as functions of wheel slip \cite{bhandari2012abswithsurfaceprediction} or through neural networks \cite{piccinini2025nnabs}.

The majority of the \ac{sc} approaches are evaluated in simulation environments \cite{basrah2017mpcslipcontrol, Pedro2012,bhandari2012abswithsurfaceprediction,pretagostini2020surveyabs,lin2003fuzzylearningabs,piccinini2025nnabs}, while only a limited number report real-world application results \cite{petersen2003lqrabs, savitski2018robust}.

Pretagostini et al. \cite{pretagostini2020surveyabs} provide an overview and compare the advantages and disadvantages of the different approaches for \ac{sc}. They conclude that model predictive control strategies offer the greatest potential for performance but introduce a significant computational burden. In contrast, rule-based methods, while requiring extensive tuning for diverse driving conditions, are computationally efficient, robust, and extensively validated in real-world scenarios.

Control algorithms for yaw stability include \ac{mpc} \cite{Parra2022nmpcesc, hashemi2020mpcesc, Li2022mpcesc}, sliding mode control \cite{zhang2020steeringesc}, \ac{lqr} \cite{Wang2018lqresc}, backstepping \cite{Zhou2010slidingmodebacksteppingesc}, and \ac{pid}-control \cite{Lenzo2021sisoyawmomentcontrol}. Typically these control algorithms are designed based on a single-track model \cite{zhang2020steeringesc, Wang2018lqresc} or a double-track model \cite{tahouni2019nonlinearcontrolesc, Parra2022nmpcesc, hashemi2020mpcesc, Li2022mpcesc, Zhou2010slidingmodebacksteppingesc}, that is combined with a linear tire \cite{zhang2020steeringesc, Wang2018lqresc}, Pacejka tire \cite{tahouni2019nonlinearcontrolesc, Parra2022nmpcesc, Zhou2010slidingmodebacksteppingesc}, or LuGre combined slip tire model \cite{hashemi2020mpcesc, Li2022mpcesc}.

Yaw stability systems are typically implemented either as single-stage systems capable of directly actuating the vehicle \cite{zhang2020steeringesc, Parra2022nmpcesc, hashemi2020mpcesc, Li2022mpcesc}, or as two-stage architectures comprising a high-level controller that generates a desired yaw moment and a low-level controller that allocates this moment to the actuators \cite{tahouni2019nonlinearcontrolesc, Zhou2010slidingmodebacksteppingesc, Wang2018lqresc, Lenzo2021sisoyawmomentcontrol}. Most designs assume the availability of positive and negative torque at each wheel, often achieved through individual electric motors \cite{Parra2022nmpcesc, hashemi2020mpcesc, Li2022mpcesc, Wang2018lqresc}. Additionally, Tahouni et al. \cite{tahouni2019nonlinearcontrolesc} introduce a system utilizing an active differential at the rear axle and a brake-based \acf{esc} system at the front axle to maintain vehicle stability. Alternative approaches employ either brake-based actuation exclusively \cite{Zhou2010slidingmodebacksteppingesc} or active steering angle adaptation \cite{zhang2020steeringesc}.

Generally, yaw stability controllers are designed to track a reference yaw rate \cite{tahouni2019nonlinearcontrolesc, Parra2022nmpcesc, Zhou2010slidingmodebacksteppingesc}, and in some cases, also track a sideslip angle \cite{zhang2020steeringesc, Wang2018lqresc} or a lateral velocity reference \cite{Li2022mpcesc, hashemi2020mpcesc}. Lenzo et al. \cite{Lenzo2021sisoyawmomentcontrol} enhance stability by adjusting the yaw rate reference with the current sideslip angle. The required reference signals are typically generated using a steady-state single-track model with linear tires \cite{tahouni2019nonlinearcontrolesc, Parra2022nmpcesc, hashemi2020mpcesc, Zhou2010slidingmodebacksteppingesc}. Additionally, some methods introduce constraints, or constraint envelopes, to limit the yaw rate and sideslip angle \cite{hashemi2020mpcesc, Li2022mpcesc}. Parra et al. \cite{Parra2022nmpcesc} specifically constrain the rear axle sideslip angle to mitigate excessive oversteer.

These yaw rate stability control concepts are predominantly evaluated in simulation environments \cite{tahouni2019nonlinearcontrolesc, Parra2022nmpcesc, Li2022mpcesc, Zhou2010slidingmodebacksteppingesc, zhang2020steeringesc}, with only a limited number reporting results from a real-world application \cite{hashemi2020mpcesc, Wang2018lqresc, Lenzo2021sisoyawmomentcontrol}.

Given the objective of implementing \ac{sc} and yaw stability algorithms for an autonomous racing vehicle to ensure safety in various conditions, the focus in this work is placed on robust, well-established concepts. Approaches with low computational requirements are prioritized to guarantee real-time performance.

\section{Methodology}

\begin{figure}[ht!]
	\centering
\inputtikzfig{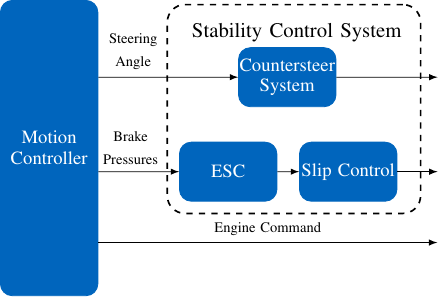}
\caption{Stability control system architecture.}
\label{fig:stability_control_architecture}
\end{figure}

Fig.~\ref{fig:stability_control_architecture} illustrates the architecture of the proposed stability control system. This system functions as an add-on to a generic motion controller, generating steering angle, individual wheel brake pressures, and an engine command. We apply the system to an autonomous race car equipped with a turbocharged internal combustion engine, where the engine is controlled solely by a throttle command. This command is not influenced, since counteracting the engine via brake pressure modulation usually offers faster overall system responses if only engine airflow can be influenced. The stability control system comprises an \ac{esc}, \ac{sc}, and \ac{cs}. The \ac{esc} and \ac{sc} systems modify the brake pressure commands, whereas the \ac{cs} modifies the steering angle command from the motion controller. The stability control system activates exclusively during critical situations, while under normal driving conditions, the motion controller retains full control over vehicle behavior.

\subsection{Electronic Stability Control}

Vehicle stability in oversteer situations is maintained by implementing an \ac{esc} system that applies brake pressure to the individual wheels of the front axle, generating a yaw moment to counteract oversteer. The brakes on the rear axle are not used to contribute a moment, since the rear tires are already saturated in an oversteer situation. Understeer situations are not addressed by the \ac{esc} system, as they are considered less critical due to slower dynamics and are assumed to be managed by the motion controller.

\begin{figure}
	\centering
	\inputtikzfig{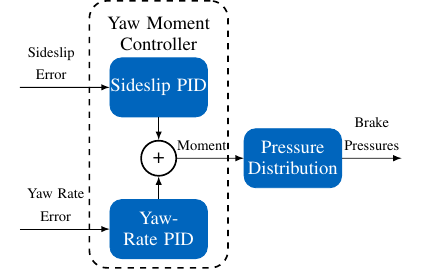}
	\caption{Structure of the \ac{esc} system.}
	\label{fig:esc_structure}
\end{figure}

As shown in Fig.~\ref{fig:esc_structure}, the \ac{esc} system is divided into an upper-level yaw-moment controller and a lower-level brake-pressure distribution strategy. The yaw moment controller regulates the yaw rate and sideslip angle errors to corresponding reference values. The reference values can be calculated from the steering angle or a target lateral acceleration. The reference generation is based on a steady-state single-track model with a linear tire assumption \cite{rajamani2006vehicle}. We calculate the yaw rate target,
\begin{equation}
	\dot{\psi}_{\mathrm{ref}} = \kappa \, v,
\end{equation}
and sideslip angle target,
\begin{equation}
	\beta_{\mathrm{ref}} = \kappa \, \left( l_r - \frac{l_f \, m \, v^2}{C_r \, l} \right),
\end{equation}
from a desired curvature value $\kappa$, vehicle velocity $v$, vehicle mass $m$, distance from center of gravity to front and rear axle $l_f$ and $l_r$ respectively, total vehicle length $l$, and rear tire cornering stiffness $C_r$. From the steering angle $\delta$, using the front tire cornering stiffness $C_f$, the curvature is calculated as
\begin{equation}
	\kappa = \delta \, \frac{C_f \, C_r \, l}{C_f \, C_r \, l^2 + m \, v^2 \, \left( l_r \, C_r - l_f \, C_f \right)},
\end{equation}
or from lateral acceleration $a_y$ as
\begin{equation}
	\label{eq:esc_target_acc}
	\kappa = \frac{a_y}{v^2}.
\end{equation}

The \ac{esc} system is activated when the filtered errors between the reference and actual values for yaw rate and sideslip angle exceed a specified threshold. Additionally, both the absolute sideslip angle and vehicle velocity must surpass a defined maximum value to trigger activation. This approach prevents unnecessary interventions during low-speed driving. The activation thresholds are dynamically adjusted based on vehicle velocity to accommodate higher absolute values of yaw rate and sideslip angle in lower speed turns with high curvature. All thresholds are manually tuned based on simulations and experimental data, aiming to minimize performance degradation from unnecessary interventions while ensuring sufficient stability in critical situations.

The yaw moment controller consists of two parallel \ac{pid}-controllers, one for yaw rate and one for sideslip angle. Before being input to the controllers, yaw rate and sideslip angle errors are low-pass filtered to minimize the influence of noise. The outputs from both controllers are combined to determine the total required yaw moment. This yaw moment is allocated to the front wheels by decreasing brake pressure on one wheel, when feasible, and increasing it equally on the opposite wheel.

\subsection{Wheel Slip Control}

The \ac{sc} is implemented as a model-free finite-state machine, as such systems demonstrate robustness and have been extensively validated in real-world applications. The presented algorithm represents an advancement over a previous version as presented in \cite{pitschi2025longcontrol}.

The \ac{sc} regulates the wheel slip by adjusting the brake pressure at each individual wheel. For acceleration, braking torque is superimposed on the engine's driving torque at the driven rear axle. 

\begin{figure}
	\centering
	\inputtikzfig{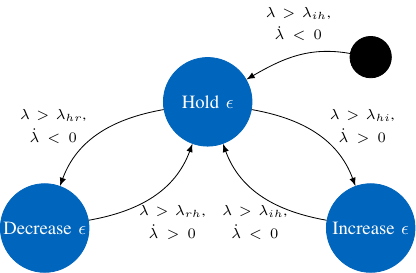}
	\caption{Schematic of the \ac{sc} finite state machine. \ac{sc} influences the brake pressure rate $\epsilon$. State transitions depend on wheel slip $\lambda$ and its derivative $\dot{\lambda}$.}
	\label{fig:abs_final_state_machine}
\end{figure}

Fig.~\ref{fig:abs_final_state_machine} presents the finite state machine logic. The state machine comprises the states hold, increase, and reduce. For each state transition, specific slip thresholds are defined. These thresholds are scaled by the slip angle at the respective wheel to account for combined tire effects. Additionally, the transitions depend on the sign of the slip ratio. The slip ratio is numerically calculated from the slip value and used to mitigate the impact of delays introduced by the brake actuator.

The \ac{sc} activates into the hold state when the slip exceeds the threshold value $\lambda_{ih}$. Upon activation, the current brake pressure is stored as a target, and subsequent brake pressure adjustments are made using a ratio $\epsilon$ of this target value. This ratio is adjusted based on the specific state within the state machine.

In the hold state, the brake pressure ratio is initially adjusted to a fixed value to enable a rapid response. Following this adjustment, the ratio remains constant. In the increase and reduce states, the pressure ratio is increased or reduced depending on the error between the actual slip and the target slip value $\lambda_{ih}$. In the increase state, the pressure ratio $\epsilon$ may be raised above $\epsilon > 1.0$ if the slip is below a defined safety threshold $\lambda_{\mathrm{save}}$. This mechanism permits corrections if the system is erroneously activated due to sensor noise or for inaccurate assumptions regarding tire potential at the onset of the braking maneuver.

In addition to the brake pressure ratio adjustments, the target value is progressively reduced during deceleration to account for the decreasing vertical tire load resulting from reduced aerodynamic downforce at lower velocities.

In the deceleration case, the \ac{sc} controls each wheel independently, whereas in the acceleration case, the rear axle is controlled as a single unit. This approach is chosen because the vehicle's locking differential counteracts wheel speed differences by distributing driving torque to the individual wheels. Therefore, controlling the individual rear-wheel brake torques during traction events would induce undesired yaw moments due to the locking differential.

\subsection{Countersteer System}

In parallel with the \ac{esc} system, a \ac{cs} is implemented that overrides the motion controller's steering angle command, in conjunction with the \ac{esc} intervention, to enhance lateral stability during critical oversteer situations. Corrections using the steering angle rather than the brake pressure command are generally preferred because they have less influence on the vehicle's longitudinal dynamics and result in less performance degradation. Additionally, for our test vehicle, the steering actuator exhibits a shorter delay of \SI{30}{\milli\second} compared to the brake actuator, which has about \SI{150}{\milli\second}, enabling a more rapid response in dynamic oversteer scenarios. Intervention thresholds for the \ac{cs} are therefore set to be more sensitive than in the \ac{esc} system.

\begin{figure}
	\centering
	\inputtikzfig{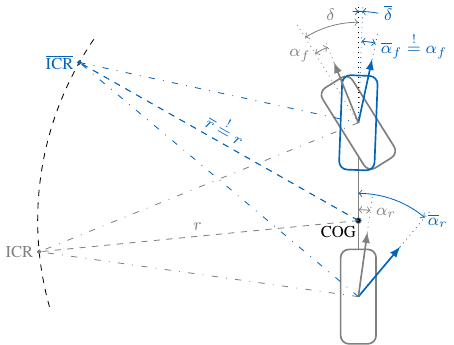}
	\caption{Countersteer system concept. The grey trajectory represents the nominal case without longitudinal load or disturbance, while the blue trajectory illustrates an oversteer situation. For both situations, the instantaneous center of rotation (ICR) is equidistant from the vehicle's center of gravity (COG). As a result, the driven radius and the yaw rate remain identical.}
	\label{fig:countersteer_architecture}
\end{figure}

In the \ac{cs}, a stationary circular motion with a curvature $\kappa = \frac{1}{r}$ is considered, as illustrated in Fig.~\ref{fig:countersteer_architecture} in grey. During this motion, we want to keep the yaw rate $\dot{\psi}$ and slip angle at the front axle $\alpha_f$ constant. Using additionally the wheelbase $l$, velocity $v$, and the rear slip angle $\alpha_r$, the required steering angle to sustain the motion can be determined by 

\begin{equation}
	\delta = \frac{\dot{\psi} \, l}{v} + \alpha_f - \alpha_r.
\end{equation}

 In an understeering vehicle configuration without longitudinal load or disturbances, the front axle exhibits a greater slip angle than the rear axle. Under these circumstances, we assume that a motion controller can maintain vehicle stability.

In contrast, when a longitudinal load is applied to the rear tires or a disturbance such as a bump on the asphalt occurs, the rear slip angle $\overline{\alpha}_r$ increases to achieve the same lateral acceleration and yaw rate, as depicted in Fig.~\ref{fig:countersteer_architecture} in blue. If the front slip angle $\overline{\alpha}_f$ should remain unchanged from the nominal case, the steering angle must be reduced to prevent an increase in yaw rate and, with this, a change in the driven curvature $\overline{\kappa} = \frac{1}{\overline{r}}$.

In a simplified abstraction assuming linear tire behavior and applying a small-angle approximation, the steering angle must be reduced by

\begin{equation}
	\Delta \delta = \overline{\alpha}_r - \alpha_f,
\end{equation}

to achieve a neutral steering condition, where $\alpha_f = \alpha_r$. A neutral or slightly understeering condition is desired, as motion controllers are generally able to maintain stability in these conditions.

Therefore, the \ac{cs} will be activated if the rear slip angle exceeds the front slip angle. In this case, the system overwrites the motion controller's steering command $\delta$ by

\begin{equation}
	\overline{\delta} = \delta - \Delta \delta.
\end{equation}

Although the \ac{cs} is designed for stationary circular motion and linear tire dynamics, our experiments demonstrate improved stability across diverse driving conditions. The feedforward estimation error due to simplifications is mitigated by a high-frequency feedback control loop. This is highlighted in the presented ablation studies, where the \ac{cs} can stabilize the vehicle even in highly nonlinear tire regions. We argue that neglecting nonlinear tire dynamics represents a strategic advantage in this case, as it enhances system robustness against external disturbances and model mismatches.
The slip angles at the front and rear axles are calculated from the individual slip angles at the corresponding right and left wheels as a weighted average, with the weights determined by the vertical load on each wheel. This approach accounts for the influence of lateral load transfer on the slip angles and provides a more accurate estimate of the required countersteer angle.
\section{Experimental Setup}

To validate the effectiveness of the proposed stability control system, we conduct both simulation studies and real-world tests.

\subsection{Simulation}

The stability control system is evaluated using different motion controllers across multiple scenarios. We use a Tube \ac{mpc} using a point mass model (\ac{pmpc}) based on Wischneswki et al. \cite{Wischnewski2023Mpc}, an adapted pure pursuit controller \cite{coulter1992ImplementationPurePursuit}, and a nonlinear \ac{mpc} based on a single-track model (\ac{nmpc}) \cite{bongard2025nmpc}. These controllers are established approaches in autonomous racing. All tracking controllers are combined with a longitudinal controller \cite{pitschi2025longcontrol} to generate brake pressure and engine commands. Unless otherwise indicated, results refer to the \ac{pmpc}. The stability control parameters are available with the source code.

For simulating vehicle dynamics, we choose a publicly available and validated implementation of a double-track model with a Pacejka Magic Formula tire model \cite{sagmeister2024sim}. The simulation incorporates simplified models of a combustion engine drivetrain and wheel suspension dynamics, and is calibrated to closely replicate the real-world behavior of the full-scale autonomous racing vehicle. To ensure reproducibility and comparability, we use a framework for deterministically executing ROS-based simulations \cite{sagmeister2026rslcpp}.

The simulation studies comprise multiple experiments. In the initial test, lateral and longitudinal acceleration references and control limits are set to conservative targets and incrementally increased in each run by a linear acceleration scaling value until an abort condition occurs. The pure pursuit controller only uses the scaled acceleration targets, while both \ac{mpc}s additionally use the provided acceleration limits. All other values are kept constant during the experiments. A run is considered successful if three laps can be completed without triggering the abort condition. The results for each configuration are reported up to the first scaling that results in an abort.

\subsection{Real-World Testing}

Experimental results stem from tests conducted during the 2025 season of the Abu Dhabi Autonomous Racing League. These tests utilized the full-scale autonomous racing vehicle Dallara EAV25, shown in Fig.~\ref{fig:hailey}. The vehicle is based on the SF23 Chassis produced by Dallara and is equipped with computational and sensing hardware as well as the required actuators for autonomous racing.

All simulation and experimental tests are conducted on the North Layout of the Yas Marina circuit in Abu Dhabi.

\section{Results}

This section shows the experimental results, starting with simulation studies, followed by the real-world experiments.

\subsection{Simulation Results}

We conduct three simulation studies to validate the effectiveness of the proposed stability control system. The first two studies are ablation studies that compare vehicle performance with and without \ac{sc} for longitudinal stability, and different stability control configurations for lateral stability. The third study evaluates the performance of the stability control system when integrated with different motion controllers.

\begin{figure}
	\centering
	\inputtikzplot{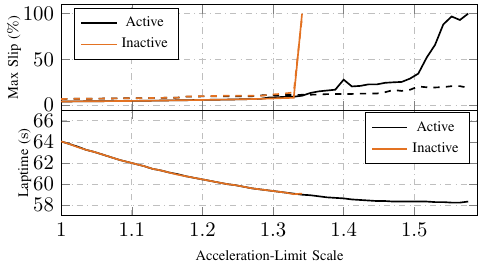}
	\caption{Simulative comparison of active and inactive \ac{sc} for increasing acceleration scaling values. Solid lines correspond to the front axle, dashed lines to the rear axle.}
	\label{fig:long_stability_sim_comparison}
\end{figure}

In the first experiment, we compare the vehicle's longitudinal stability with and without \ac{sc}. The results in Fig.~\ref{fig:long_stability_sim_comparison} present the maximum wheelslip values on the front and rear axles and laptimes during each run for increasing acceleration scaling values. The test is terminated if a tire begins to block or spin.

With \ac{sc}, the controller can handle higher reference values and achieve lower lap times without tire blocking or spinning. Furthermore, the maximum slip values increase gradually with higher scaling values, in contrast to the rapid increase observed without \ac{sc} on both the front and rear axles. This gradual increase facilitates the detection of approaching vehicle limits and enables corrective actions.

\begin{figure}
	\centering
	\inputtikzplot{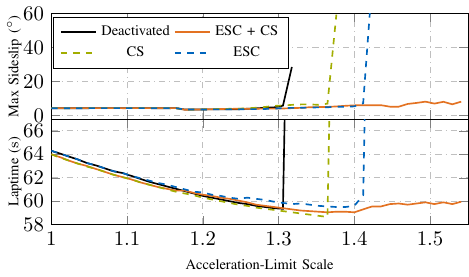}
	\caption{Simulative comparison of different stability control configurations for increasing acceleration scaling values.}
	\label{fig:lat_stability_sim_comparison}
\end{figure}

The second experiment evaluates the performance of various lateral stability control configurations. No stability control, only \ac{esc}, only \ac{cs}, and both \ac{esc} and \ac{cs} are considered. To promote oversteering, the rear lateral simulation grip is reduced by \SI{20}{\percent}. The test is terminated if the maximum sideslip angle exceeds \SI{20}{\degree} or if the maximum lateral error from the reference trajectory exceeds \SI{3}{\metre}. Fig.~\ref{fig:lat_stability_sim_comparison} presents the maximum sideslip angle experienced by the vehicle and the laptimes as the acceleration limits are increased.

The configuration without stability control yields the lowest acceleration scaling value. In contrast, all other configurations stabilize the vehicle at higher scaling values. The configurations with only \ac{cs} and with both systems active achieve reduced lap times. However, when only \ac{cs} is enabled or when both systems are disabled, the acceleration scaling limit associated with the lowest lap time corresponds to the highest stable reference. Consequently, any further increase in the reference value leads to abrupt instability, which is difficult to predict during testing. The configuration using only \ac{cs} achieves the lowest lap times since it does not reduce the vehicle's velocity with brake interventions and should therefore be preferred from a performance perspective. However, the absence of a predictable stability limit makes it unsafe for application on the real vehicle.

In contrast, when both systems are enabled or only \ac{esc} is enabled, the scaling that yields the minimum lap time is lower than the maximum stable scaling. This allows approaching the performance limits more safely. However, with only \ac{esc} enabled, the lap time remains higher than when both systems are active. For a scaling value of $GG = 1.3$, using only the \ac{esc} results in a lap-time difference of \SI{0.15}{\second} compared to the configuration without stability control. The configuration with both \ac{esc} and \ac{cs} can reduce this difference to \SI{0.05}{\second}, indicating that the \ac{cs} contributes to improved vehicle performance and stability.

\begin{figure}
	\centering
	\inputtikzplot{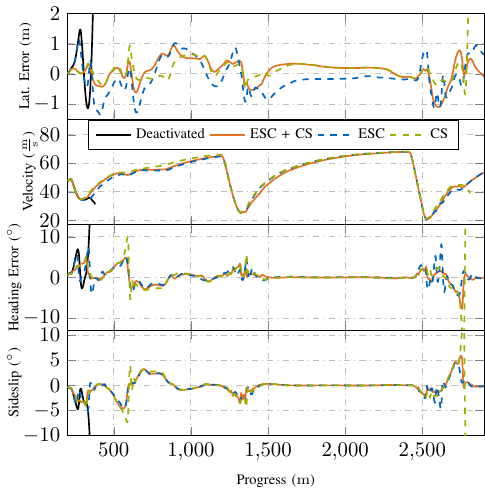}
	\caption{Simulative comparison of different stability control configurations for the first lap at an acceleration scaling value of $GG = 1.4$.}
	\label{fig:lat_stability_sim_comparison_timeseries}
\end{figure}

Fig.~\ref{fig:lat_stability_sim_comparison_timeseries} shows the timeseries of the lateral error, velocity, heading error, and sideslip angle for the run with an acceleration Scale of $GG = 1.4$ for all four configurations.

The configuration without lateral stability control spins in the first turn, resulting in the test being aborted. While the configuration with only the \ac{cs} completes the majority of the lap, it ultimately spins in the final turn. In contrast, the configurations utilizing only \ac{esc} or both stability systems maintain vehicle stability throughout the entire run, with consistently lower sideslip angles. Notably, the configuration with both \ac{esc} and \ac{cs} achieves the lowest heading and lateral position errors.

The \ac{cs}-only configuration maintains the highest velocity, but cannot keep the vehicle stable throughout the run. Additionally, the configuration with both \ac{esc} and \ac{cs} leads to a higher velocity than the configuration with only \ac{esc} activated. This shows that the combination of \ac{esc} and \ac{cs} positively complements each other in terms of vehicle stability and performance.

\begin{figure}
	\centering
	\inputtikzplot{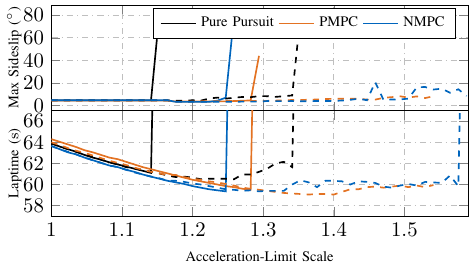}
	\caption{Simulative comparison of stability control for different motion controllers for increasing acceleration scaling values. Solid lines correspond to the case without stability control, dashed lines correspond to the case with stability control using both \ac{esc} and \ac{cs}.}
	\label{fig:lat_stability_sim_comparison_controllers}
\end{figure}

The final simulation study evaluates the performance of the stability control system when coupled with different motion controllers. The \ac{esc} targets are calculated using Eq.~\ref{eq:esc_target_acc} for all motion controllers. Fig. ~\ref{fig:lat_stability_sim_comparison_controllers} presents the maximum sideslip angles and laptimes for increasing acceleration scale values for each motion controller, both with and without stability control activated. Consistent with the lateral stability ablation study, rear lateral tire grip is reduced by \SI{20}{\percent}, and tests are terminated if the maximum sideslip angle exceeds \SI{20}{\degree} or if the maximum lateral error exceeds \SI{3}{\metre}.

The stability control system allows all motion controllers to operate with higher acceleration scaling values while maintaining vehicle stability. When stability control is implemented, both the pure pursuit controller and \ac{pmpc} achieve reduced lap times. In contrast, without stability control, the vehicle spins before a visible minimum in lap time has been reached. For the \ac{nmpc}, the same minimum lap time is achieved with and without stability control.

However, without stability control, the maximum scaling value that permits stable driving coincides with the minimum lap time for all controllers.  As a result, it is not possible to predict whether increasing the limits is safe and improves lap time or causes a loss of control. In contrast, with the stability control system, the maximum achievable scaling value is higher than the value corresponding to the minimum lap time. Consequently, the stability control system enables a robust exploration of the performance limit of the corresponding controller, as the vehicle does not spin immediately upon reaching the lap-time-optimal reference value, which is not known beforehand.

The \ac{pmpc} with stability control could operate the vehicle at higher scaling values than shown without spinning, but the tests were terminated because the lateral error threshold was exceeded due to understeering.

\subsection{Real-World Results}

In addition to validation through simulation studies, we conducted real-world tests during the 2025 season of the Abu Dhabi Autonomous Racing League. We carried out these tests over 4.5 weeks in total, with an overall track time of about 40 hours, without any crashes, spins, or damaged tires. The best laptime achieved was within a second of former formula one driver Daniil Kvyat on the same track with the same type of vehicle, less than two hours before our record lap. The maximum accelerations during these tests reached up to \SI{29.2}{\metre\per\second\squared} longitudinal and \SI{29.5}{\metre\per\second\squared} lateral acceleration. This shows that the stability control system can reliably maintain the vehicle's safety at a high performance level.

We present the results of two situations in which the stability control system intervened during the tests. The first situation is one where the vehicle entered an oversteering state, and the stability control system intervened to maintain safety. The second situation is a case of regular driving in which countersteering is applied to maintain stability without performance degradation.

\begin{figure}
	\centering
	\inputtikzplot{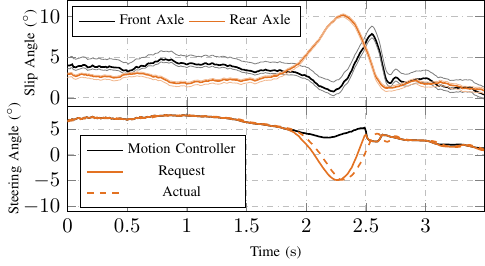}
	\caption{Countersteer intervention during a real world test. The semi-transparent lines correspond to the slip angles on the left and right side of the vehicle, while the solid line corresponds to the weighted slip angle.}
	\label{fig:countersteer_faulty_motion_controller}
\end{figure}

\begin{figure}
	\centering
	\inputtikzplot{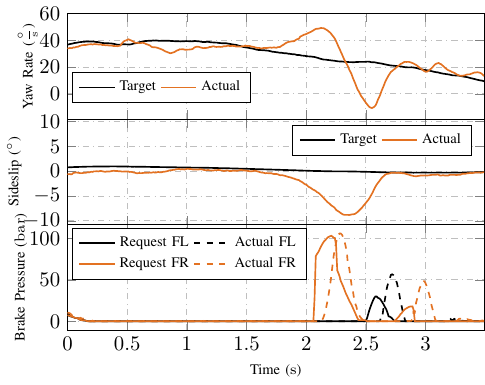}
	\caption{\ac{esc} intervention during a real world test.}
	\label{fig:esc_faulty_motion_controller}
\end{figure}

Fig.~\ref {fig:countersteer_faulty_motion_controller} and Fig.~\ref{fig:esc_faulty_motion_controller} show the intervention of the \ac{cs} and \ac{esc}, respectively. Due to excessive throttle during acceleration out of a left turn, the vehicle starts to spin, which is detected by the stability control system. It triggers an intervention from both \ac{esc} and \ac{cs} to stabilize the vehicle.

The slip angle at the rear axle increases continuously relative to that at the front axle. The steering angle is therefore decreased by the \ac{cs} to counteract the vehicle's turning in. Additionally, the ESC system is activated during the situation, when the sideslip and yaw rate error thresholds are exceeded. The \ac{esc} initially applies brake pressure to the front outer wheel to reduce the yaw rate and turn the vehicle back toward the reference. After the sideslip angle is reduced, the system applies brake pressure to the front inner wheel, followed by a command to the outer wheel to stabilize the vehicle and return the yaw rate to the reference value.

Without the stability control system's intervention, the vehicle's yaw motion would have accelerated, potentially leading to a spin or a crash.

\begin{figure}
	\centering
	\inputtikzplot{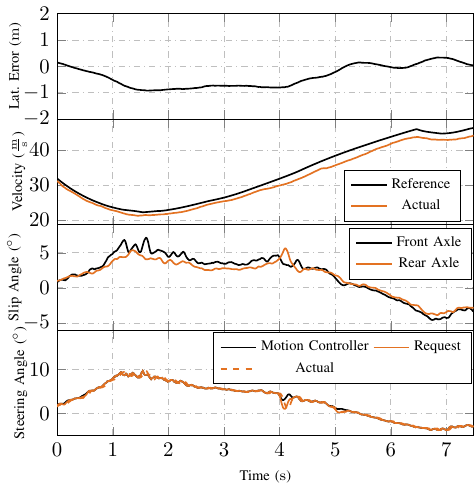}
	\caption{Countersteer intervention during fastest lap of testing on the Yas Marina Circuit.}
	\label{fig:countersteer_regular_operation}
\end{figure}

As a second example, we show in Fig.~\ref{fig:countersteer_regular_operation} the intervention of the \ac{cs} during regular operation in the final event on the Yas Marina Circuit. In this situation, the vehicle is driving through a combination of two left turns at low to mid speeds. In the second turn, the rear slip angle increases twice above the front slip angle at $t = \SI{4}{\second}$ and $t = \SI{5}{\second}$. The \ac{cs} responds by reducing the steering angle to maintain stability. With this intervention, the vehicle maintains a stable trajectory, and the lateral error remains low. At the same time, no brake intervention by the \ac{esc} system or velocity reduction by the motion controller is necessary, and the vehicle can maintain high acceleration throughout the turn.

\section{Conclusion}
\label{sec:conclusion}

We presented a robust stability control approach comprising an \ac{esc} system, a \ac{sc}, and a \ac{cs}. The effectiveness of this stability control system was demonstrated through both simulation and experimental validation. Our results indicate that the stability control system enhances the robustness of motion controllers against faulty control commands and external disturbances. Furthermore, the operating range of a motion controller is expanded, allowing exploration at the limits of its capabilities without loss of control, thereby increasing safety and predictability when approaching its performance limits. Nevertheless, the overall system performance is primarily determined by the underlying motion controller, and the stability control system offers limited improvement if the motion controller is less performant. In future work, the system can be extended with a Torque Vectoring System to regulate lateral dynamics continuously during both understeering and oversteering, thereby enhancing vehicle agility. 

\section*{Acknowledgment}
Author contributions: Phillip Pitschi, as the first author, designed the structure of the article and contributed essentially to the development and implementation of the concept. S. Sagmeister contributed to the paper writing and the concept's design and implementation. Frederik Werner contributed to the concept's design, verification and data analysis.
Boris Lohmann and Markus Lienkamp made an essential contribution to the concept of the research project.
AI tools (GPT5.2, Gemini 3 Pro) were used exclusively for editorial, code, and documentation refinement of author-provided input.

%
%

\end{document}